%% file: main_traffic.tex
\documentclass[conference,10pt]{IEEEtran}
\IEEEoverridecommandlockouts

\usepackage{amsmath,amsfonts,amssymb,amsthm}
\usepackage{graphicx}
\usepackage{booktabs}
\usepackage{multirow}
\newcommand{\norm}[1]{\lVert #1 \rVert}
\usepackage[hidelinks]{hyperref}
\usepackage{xcolor}
\usepackage[capitalize,noabbrev]{cleveref}
\usepackage[numbers]{natbib}

\newtheorem{proposition}{Proposition}
\newtheorem{theorem}{Theorem}

\graphicspath{{figures/}}

\title{When Does Physics Beat Data? Data-Efficient Traffic State
Estimation with Amortized PINNs}

\author{\IEEEauthorblockN{Anna Lazareva\IEEEauthorrefmark{1}\IEEEauthorrefmark{2} and
Alexander Tarakanov\IEEEauthorrefmark{1}\IEEEauthorrefmark{3}}
\IEEEauthorblockA{\IEEEauthorrefmark{1}HSE University, Moscow, Russia}
\IEEEauthorblockA{\IEEEauthorrefmark{2}Moscow Institute of Physics and Technology (MIPT), Russia}
\IEEEauthorblockA{\IEEEauthorrefmark{3}AI VK, Moscow, Russia}
\IEEEauthorblockA{lazareva.anna@phystech.edu, annelzrv@gmail.com, atarakanov@hse.ru}}

\newcommand{\runinhead}[1]{\par\textit{#1:}\ }

\begin{document}
\maketitle
\begingroup\renewcommand\thefootnote{}\footnotetext{Accepted at the SSTDM workshop, IEEE ICDM 2026 (ICDMW). \copyright\ 2026 IEEE. Personal use of this material is permitted. Permission from IEEE must be obtained for all other uses, in any current or future media, including reprinting/republishing this material for advertising or promotional purposes, creating new collective works, for resale or redistribution to servers or lists, or reuse of any copyrighted component of this work in other works.}\addtocounter{footnote}{-1}\endgroup

\input{sections/abstract_traffic.tex}
\input{sections/introduction_traffic.tex}
\input{sections/related_work_traffic.tex}
\input{sections/method_traffic.tex}
\input{sections/experiments_traffic.tex}
\input{sections/results_traffic.tex}
\input{sections/theory_traffic.tex}
\input{sections/discussion_traffic.tex}


\bibliographystyle{IEEEtran}
\bibliography{main_traffic}

\end{document}

%% file: sections/abstract_traffic.tex
\begin{abstract}
Traffic state estimation (TSE) reconstructs traffic density at
unmonitored locations from a sparse set of sensors. Data-driven models
work well when sensors are dense but degrade sharply as coverage thins.
We show that imposing the Lighthill--Whitham--Richards conservation law
as a physics-informed prior makes TSE more data-efficient. On four real
corridors (NGSIM, I-80, PeMS04, PeMS08), physics-informed estimators
generalise to held-out sensors where data-driven models, including
classical and deep kriging, overfit the observed ones. On the two
shock-rich corridors the effect is a clear crossover: as more sensors
are held out, data-driven models collapse while the physics-informed
estimator degrades gracefully, leading the no-physics network by
$\approx\!26$--$38\%$ at $85$--$90\%$ held out and the strongest classical
kriging by $\approx\!10$--$21\%$ at $90\%$. A matched smoothness
prior does not reproduce the gain, and an extended Kalman filter on the
same LWR model is outperformed on the shock corridors, so the advantage comes from how the
conservation law is imposed. Finally, one loss-conditional PINN removes
the per-corridor loss-weight search, matching a 20-point search at
$\sim\!5\times$ lower cost.
\end{abstract}

%% file: sections/introduction_traffic.tex
\section{Introduction}
\label{sec:intro}

Traffic state estimation (TSE) reconstructs the density and speed of
traffic along a road from a sparse set of fixed sensors. It feeds ramp
metering, incident detection and routing. Sensors are expensive, so real
corridors are instrumented sparsely and unevenly, and the important
question is what traffic does \emph{between} the sensors. This is a
spatiotemporal kriging problem, and it becomes harder as coverage thins.

Data-driven estimators, including modern spatiotemporal graph networks,
work well when sensors are dense because they learn spatial correlation
from the observed signals. That correlation must be estimated from data,
and with fewer sensors there is less data to estimate it from, so these
methods degrade sharply under sparse sensing, the regime that matters
for deployment.

Traffic flow obeys a conservation law, the Lighthill--Whitham--Richards
(LWR) model, which constrains the density over the whole road, not only
at the sensors. Embedding this law in a network's training objective, as
physics-informed neural networks (PINNs) do, gives a prior that holds
where no data does. Physics-informed TSE is not new: Shi et
al.~\citep{shi2021lwr} proposed a PINN on the LWR model and observed that
the prior helps as detectors thin. We turn that observation into a
quantified and explained result.

\runinhead{Contributions}
\begin{itemize}
\item \textbf{Physics is data-efficient (the main result).} On four real
corridors (NGSIM, I-80, PeMS04, PeMS08), physics-informed estimators
generalise to unmonitored sensors where no-physics networks, Gaussian-process kriging and deep inductive kriging (IGNNK) collapse as coverage
thins, and even the most robust baseline ($k$NN kriging) is overtaken at
the sparsest coverage. On the two shock-rich corridors (NGSIM and I-80)
this is a crossover that replicates: data-driven methods win when dense,
physics wins when sparse, by $\approx\!26$--$38\%$ relative to the
no-physics network and $\approx\!7$--$21\%$ against the strongest
classical kriging at $85$--$90\%$ held out. On PeMS08 physics leads the
strongest classical baseline at every coverage; near-free-flow PeMS04 is closer (Table~\ref{tab:sweep}). Where shock structure is present it
is also ahead of an EKF on the same LWR model.
\item \textbf{The gain is the conservation law, not smoothing.} A matched
generic-smoothness prior is worse than a no-physics network.
\item \textbf{No weight search via loss-conditioning.} A loss-conditional
PINN (LC-PINN) trains once over a prior on the data-vs-physics loss
weights, amortising the per-corridor weight search ($\sim\!5\times$
cheaper than a measured search, $18$--$58\times$ vs.\ an exhaustive grid)
and, on the PeMS corridors, slightly improving on a tuned fixed-weight
PINN.
\item \textbf{A theoretical account:} an a-posteriori error certificate at
\emph{unmonitored} locations that data-driven estimators lack.
\end{itemize}
The contribution is the phenomenon and its explanation, not a new
architecture. Held-out TSE at unmonitored locations is hard: our result
is the \emph{relative} physics-vs-data-driven gap, and at a single fixed
weighting a tuned PINN remains slightly more accurate than the amortised
LC-PINN (\S\ref{sec:results:darcy}, \S\ref{sec:discussion}).

%% file: sections/related_work_traffic.tex
\section{Related Work}
\label{sec:related}

\runinhead{Traffic state estimation and physics-informed learning}
Classical TSE assumes a traffic-flow model, typically the first-order
LWR conservation law \citep{lighthill1955kinematic,richards1956shockwaves},
and estimates the state by filtering or variational data assimilation;
the extended Kalman filter (EKF) is the standard tool, with measurements
aggregated by Edie's definitions \citep{edie1963}. Physics-informed deep
learning instead puts the flow model into a neural network's loss. The
closest prior work is Shi et al.\ \citep{shi2021lwr}, who introduce
PINN-based TSE on the same LWR model, show that the physical prior beats
NN, LSTM and EKF estimators as detectors become sparse, and name the
choice of data-vs-physics loss weights as an open problem (a companion
paper treats the second-order ARZ model \citep{shi2021pidl}). We build on
this work. We condition the network on the loss weights instead of tuning
them per corridor; we compare against deep spatiotemporal kriging and an
EKF on the same LWR model; we isolate the gain with a matched-smoothness
control; and we give the effect a theoretical account.

\runinhead{Data-driven traffic models}
Spatiotemporal graph networks such as DCRNN \citep{li2018dcrnn} and ASTGCN
\citep{guo2019astgcn}, trained on detector archives such as Caltrans PeMS
\citep{chen2001pems}, target \emph{forecasting}: the future state at
sensors observed during training. Held-out-sensor TSE is the
complementary \emph{kriging} problem, the present state where no sensor
exists, so we compare against kriging models built for that task.

\runinhead{Spatiotemporal kriging and imputation}
Gaussian-process kriging and nearest-neighbour interpolation reconstruct
unobserved locations from spatial correlation. Deep inductive methods
learn this correlation: IGNNK \citep{wu2021ignnk} uses inductive graph
message passing for virtual sensors, and GRIN \citep{cini2022grin} uses a
bidirectional recurrent graph network. We use GP, $k$NN and IGNNK on
every corridor.

\runinhead{Amortised conditional inference}
Conditioning a network on a quantity normally fixed by hand, and training
over a prior on it, comes from loss-conditional training in generative
modelling \citep{dosovitskiy2020}. The loss-conditional PINN
(LC-PINN)~\citep{lazareva2026lcpinn} extended it to parametric PDEs; we
adopt it for TSE to amortise loss-weight selection. Unlike the generative
case, the trade-offs here share a common minimiser
(Proposition~\ref{prop:invariance}), so conditioning serves robustness
and amortisation, not navigation of a preference frontier
(Section~\ref{sec:method:inference}).

%% file: sections/method_traffic.tex
\section{Method}
\label{sec:method}

We cast traffic state estimation (TSE) as a held-out-sensor
reconstruction problem and solve it with a physics-informed neural
network whose physics is the LWR conservation law. To remove the
loss-weight tuning that physics-informed estimators normally require,
we condition the network on the loss weights themselves and train a
single network over a prior on them, recovering tuned-quality estimates
without a per-corridor weight search.

\subsection{Traffic state estimation as sensor kriging}
\label{sec:method:tse}

A corridor is a one-dimensional road segment $x \in [0, L]$ observed
over a window $t \in [0, T]$. Its macroscopic state is the vehicle
density $\rho(x, t)$ (equivalently the normalized occupancy), from
which speed and flow follow through a fundamental diagram (FD). We observe
$\rho$ only at a sparse set of sensor locations: an \emph{observed}
set $\mathcal{S}_\mathrm{obs} \subset [0, L]$ and a disjoint
\emph{held-out} set $\mathcal{S}_\mathrm{test}$ of locations with no
sensor. The estimation task is to reconstruct $\rho$ on the full
space-time domain, and we report the relative-$L^2$ error on the
held-out sensors,
\begin{equation}
\label{eq:heldout-metric}
\mathrm{rel\text{-}L^2}_\mathrm{test}
 = \frac{\big\lVert \widehat{\rho} - \rho \big\rVert_{L^2(\mathcal{S}_\mathrm{test} \times [0,T])}}
        {\big\lVert \rho \big\rVert_{L^2(\mathcal{S}_\mathrm{test} \times [0,T])}} ,
\end{equation}
which measures generalisation to \emph{unmonitored} locations rather
than fit at observed ones. This is the spatiotemporal-kriging (sensor
imputation) setting, distinct from forecasting: we estimate the present
state where no sensor exists, not the future state where one does.

\subsection{LWR physics}
\label{sec:method:lwr}

The Lighthill--Whitham--Richards model
\citep{lighthill1955kinematic,richards1956shockwaves} states
conservation of vehicles,
\begin{equation}
\label{eq:lwr}
\partial_t \rho + \partial_x Q(\rho) = 0,
\qquad Q(\rho) = \rho\, V(\rho),
\end{equation}
where the flow $Q$ is a function of density alone through a fundamental
diagram $V(\rho)$. We use the Greenshields relation
$V(\rho) = v_f\,(1 - \rho/\rho_{\max})$, with free-flow speed $v_f$ and
jam density $\rho_{\max}$ calibrated per corridor; a triangular diagram
is evaluated as an ablation. Because $Q$ is strictly concave (hence
nonlinear), the entropy
solution of \eqref{eq:lwr} develops shocks --- the backward-propagating
stop-and-go fronts of congested traffic --- which a conservation-law
prior reconstructs but a generic smoothness prior cannot
(Section~\ref{sec:results:control}). For a smooth, entropy-admissible
residual we add a small viscous term $\varepsilon\,\partial_{xx}\rho$
($\varepsilon{=}10^{-2}$): the diffusively-corrected LWR also used for
physics-informed TSE by Shi et al.~\citep{shi2021lwr}.

\subsection{Physics-informed loss and loss-conditioning}
\label{sec:method:lcpinn}

Let $u_\theta(x, t)$ approximate the (normalized) density. A standard
physics-informed estimator minimises a weighted sum of a data term on
the observed sensors, the LWR residual on interior collocation points,
and initial/boundary terms:
\begin{equation}
\label{eq:pinn-loss}
\mathcal{L}(\theta; \lambda)
 = \lambda_\mathrm{d}\,\mathcal{L}_\mathrm{data}
 + \lambda_\mathrm{r}\,\mathcal{L}_\mathrm{res}
 + \lambda_\mathrm{i}\,\mathcal{L}_\mathrm{ic}
 + \lambda_\mathrm{b}\,\mathcal{L}_\mathrm{bc},
\end{equation}
with $\lambda = (\lambda_\mathrm{d}, \lambda_\mathrm{r},
\lambda_\mathrm{i}, \lambda_\mathrm{b})$ on the simplex
$\Delta^{3}$. The residual term is non-dimensionalised by the
characteristic wave speed $C = v_f\, L_t / L_x$ (here $C \approx 20$)
so that its gradients do not swamp the data term:
$\mathcal{L}_\mathrm{res} = \lVert \partial_t u_\theta + C^{-1}
\partial_x Q(u_\theta) \rVert^2$.

\begin{figure*}[t]
\centering
\includegraphics[width=0.7\textwidth]{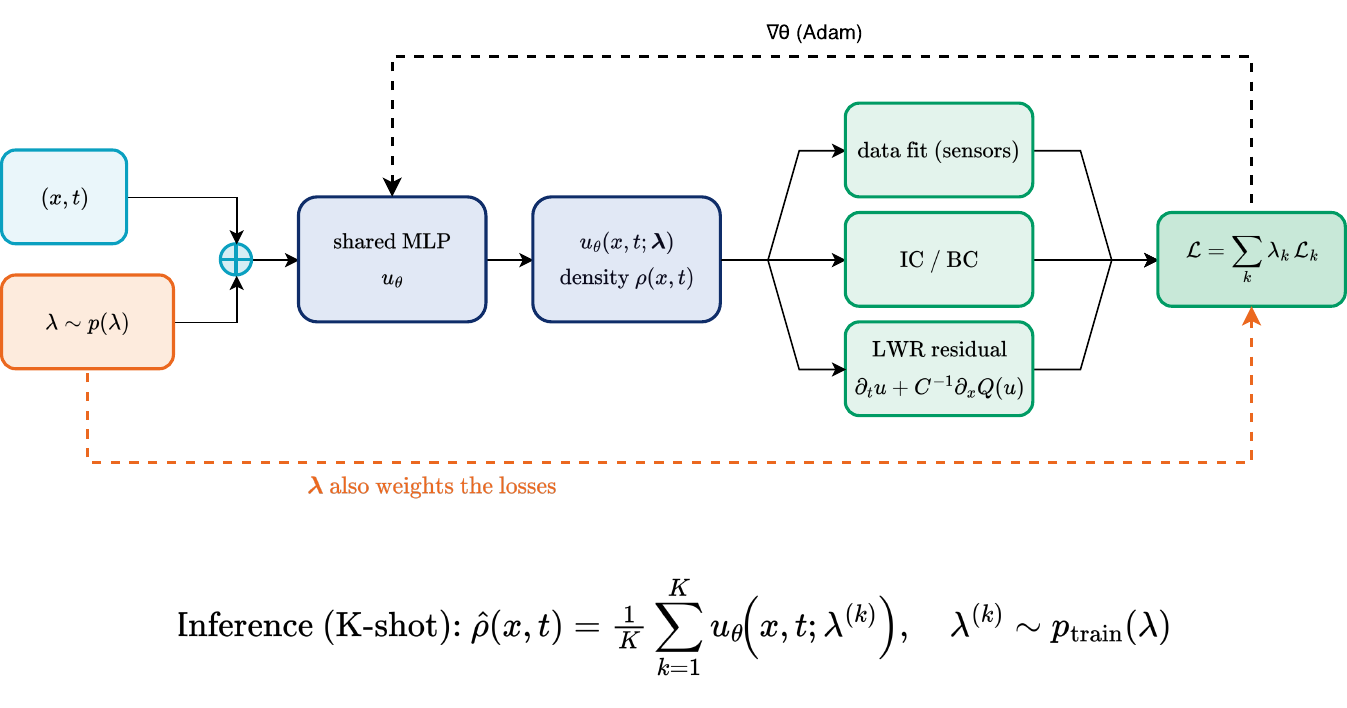}
\caption{LC-PINN for traffic state estimation. Loss weights
$\boldsymbol\lambda\sim p(\boldsymbol\lambda)$ are concatenated to the
inputs $(x,t)$ ($\oplus$), so one network
$u_\theta(x,t;\boldsymbol\lambda)$ represents every weighting; the same
$\boldsymbol\lambda$ weights the data, LWR-residual and
initial/boundary losses. At inference $\boldsymbol\lambda$ can be $K$-shot averaged
(Proposition~\ref{prop:invariance}); we report centre-$\boldsymbol\lambda$ estimates.}
\label{fig:arch}
\end{figure*}

The estimate is notoriously sensitive to $\lambda$: too much residual
weight ignores the sensors, too little discards the physics. The
standard remedy is a per-corridor weight search. Instead we use the
loss-conditional PINN (LC-PINN)~\citep{lazareva2026lcpinn}, which adapts
the loss-conditional construction \citep{dosovitskiy2020} to the
PDE-residual setting: $\lambda$ is supplied to the network as an input
and one network is trained to minimise
the expected loss over a prior $p(\lambda)$,
\begin{equation}
\label{eq:lc-objective}
J(\theta) = \mathbb{E}_{\lambda \sim p(\lambda)}\big[\,
  \mathcal{L}\big(\theta; \lambda, u_\theta(\cdot, \log\lambda)\big) \big],
\end{equation}
estimated at each step from $n_\lambda$ samples of $\lambda$. The
weight vector enters by input-level concatenation of $\log\lambda$;
because $\lambda$ reweights the objective rather than altering the PDE,
input concatenation is the appropriate injection (hidden-layer
modulation gives no consistent gain in this regime). The resulting
network is the LC-PINN for traffic state estimation
(Figure~\ref{fig:arch}). It amortises the
weight search: one training run covers the whole weight frontier, and
no $\lambda$ need be chosen before training. Because the held-out
optimum is essentially $\lambda$-invariant
(Proposition~\ref{prop:invariance}), the estimate is insensitive to the
exact prior $p(\lambda)$ --- we use a log-uniform prior on the simplex
--- and $K$-shot averaging over the trained region only reduces
variance, not bias; we therefore do not tune the prior per corridor.

\subsection{Inference and the role of \texorpdfstring{$\lambda$}{lambda}}
\label{sec:method:inference}

At inference there is no canonical $\lambda$, so we average the network
over $K$ samples from the prior,
\begin{equation}
\label{eq:k-shot}
\widehat{u}_K(x, t) = \tfrac{1}{K} \sum_{k=1}^{K}
  u_\theta\!\big(x, t, \log\lambda^{(k)}\big),
  \qquad \lambda^{(k)} \sim p(\lambda),
\end{equation}
which we call $K$-shot averaging. Its justification is an invariance
property of the optimum.

\begin{proposition}[Loss-weight invariance at the optimum]
\label{prop:invariance}
Let $p(\lambda)$ have full support on $\Delta^{3}$ and suppose the
network class contains, for each $\lambda$, a minimiser of
$\mathcal{L}(\cdot\,;\lambda)$ that drives every loss term to zero
(an idealisation for noise-free, model-matched data; on real corridors
it holds only approximately, hence the residual $\lambda$-dependence
noted below). Then
for any $\theta^\star \in \arg\min_\theta J(\theta)$ and
$p$-almost every $\lambda$, the section $u_{\theta^\star}(\cdot, \log\lambda)$
is a residual minimiser of $\mathcal{L}(\cdot\,;\lambda)$; in particular
it is independent of $\lambda$ on the support of $p$.
\end{proposition}

Proposition~\ref{prop:invariance} (established for LC-PINN
by~\citep{lazareva2026lcpinn})\footnote{Proofs, full-field errors, full tuned-baseline numbers, the sensor-budget, noise and $K$-shot curves, and the exact sensor splits are available from the authors.} frames $\lambda$ as a robustness
knob rather than a preference selector: at the population optimum the
estimate does not depend on which $\lambda$ is queried. At finite
training a small residual $\lambda$-dependence remains, and
\eqref{eq:k-shot} acts as a Monte~Carlo ensemble that reduces its
standard deviation at rate $\mathcal{O}(K^{-1/2})$ while leaving the
$\lambda$-invariant mean unchanged. Consistent with this, $K$-shot
averaging does not improve mean accuracy over evaluating at the prior
centre; we therefore report centre-$\lambda$ estimates and use $K$-shot
only to quantify residual non-invariance.

\subsection{Baselines}
\label{sec:method:baselines}

We compare against the natural alternatives for held-out-sensor TSE,
grouped by how much physics they use. \emph{Physics:} the tuned
fixed-weight LWR-PINN (the strongest single-weighting estimator, whose
weights are themselves found from the LC-PINN frontier). \emph{No
physics:} the same network with the residual weight set to zero
(``data-only''), and a generic Laplacian-smoothness prior that
penalises curvature without the conservation law --- the control that
isolates whether it is the \emph{physics} that helps rather than any
regulariser. \emph{Data-driven kriging:} Gaussian-process spatiotemporal
kriging and $k$-nearest-neighbour interpolation (classical), and the
deep inductive-kriging model IGNNK \citep{wu2021ignnk}. Forecasting models such as DCRNN
\citep{li2018dcrnn} and ASTGCN \citep{guo2019astgcn} predict the future
state at \emph{monitored} sensors and are not estimators for
unmonitored locations; we therefore position them as related work
rather than as baselines, to avoid a task mismatch.

%% file: sections/experiments_traffic.tex
\section{Experimental Setup}
\label{sec:experiments}

\runinhead{Corridors}
We evaluate on four real freeway corridors.\footnote{All datasets are
public: NGSIM (US DOT/FHWA), \url{https://ops.fhwa.dot.gov/trafficanalysistools/ngsim.htm};
the PeMS04/PeMS08 benchmark subsets released with~\citep{guo2019astgcn},
\url{https://github.com/Davidham3/ASTGCN}; underlying Caltrans PeMS
system~\citep{chen2001pems}, \url{https://pems.dot.ca.gov}.} \textbf{NGSIM US-101} and
\textbf{NGSIM I-80} (US DOT/FHWA) provide vehicle trajectories, which we
aggregate into a space-time density field by Edie's definitions
\citep{edie1963}; both are congested-peak recordings that exhibit
genuine backward-propagating stop-and-go waves (I-80 is the
more congested of the two). \textbf{PeMS04} and \textbf{PeMS08} are
Caltrans loop-detector corridors \citep{chen2001pems} that report
\emph{both} flow and speed, so density follows directly as
$\rho = q/v$ without assuming a fundamental diagram; we select the
longest contiguous sensor chain and a congested window. All four
corridors thus carry
\emph{independently-measured} density (we deliberately exclude
speed-only corridors, where density would have to be inferred from the
same fundamental diagram the physics enforces). Each corridor is
discretised on a space-time grid; the macroscopic state is the
(normalised) density $\rho(x,t)$.

\runinhead{Held-out-sensor protocol}
Sensors are interior grid locations. For a held-out fraction
$f \in \{0.30, 0.50, 0.70, 0.85, 0.90\}$ we randomly designate $f$ of
the interior sensors as held-out (test) and keep the rest as observed
(train), retaining the endpoints; the split is reseeded per random
seed. All methods see the same observed sensors and are scored by the
held-out rel-$L^2$ of \eqref{eq:heldout-metric}. We emphasise that this
metric is evaluated \emph{only at the held-out sensors} --- locations
with no observation at any time --- which is strictly harder than the
full-field error (averaged over observed cells as well) reported in much
of the TSE literature \citep{shi2021lwr}; our absolute numbers are
correspondingly larger and are not directly comparable to full-field
results. We report means over
five seeds for every method, with bootstrap $95\%$ CIs in
Figure~\ref{fig:crossover}.

\runinhead{Models and training}
All PINNs use a $4\times64$ tanh MLP. The fixed-weight LWR-PINN and the
no-physics network minimise \eqref{eq:pinn-loss} with the indicated
weights; LC-PINN concatenates $\log\lambda$ and trains over a log-uniform
prior on the simplex with $n_\lambda{=}4$ samples per step. Training is
Adam ($10^{-3}$, cosine schedule, gradient clipping, $25$k steps); the
residual is non-dimensionalised by $C{=}v_f L_t/L_x$. The fundamental
diagram is Greenshields (triangular as an ablation), calibrated as two
global scalars ($v_f,\rho_{\max}$) from the aggregate flow--density
scatter; it encodes the corridor's bulk regime, not per-location state,
so it transmits no held-out sensor \emph{values} to the estimator.
Classical kriging
(GP, $k$NN) and the deep inductive-kriging baselines (IGNNK) are fit on the observed sensors and predict the held-out ones;
GP/$k$NN are training-free at inference and IGNNK uses $3$k epochs.
All data-driven baselines use standard settings ($k$NN $k{=}6$, GP with
an RBF kernel); like the physics methods, they receive no per-corridor
hyperparameter search, so no estimator is advantaged by tuning effort.
As a classical model-based counterpart we also run an extended Kalman
filter (EKF) data-assimilation baseline on the \emph{same} calibrated
Greenshields LWR dynamics (Godunov discretisation, ensemble formulation
for robustness to shocks), observing the train sensors at every time
step and estimating the held-out ones; it uses a single fixed
process/observation-noise setting across all corridors, so it is, like
LC-PINN, free of per-corridor tuning.
The external PDEBench 2-D Darcy benchmark uses the same conditioning
applied to a permeability-parameterised elliptic problem.

\runinhead{Reproducibility}
After Edie/loop aggregation the corridors are NGSIM US-101
($56\times180$ cells, $40$\,ft$\times5$\,s, 8 lanes), NGSIM I-80
($45\times394$, 7 lanes, $\sim$30-min congested window), and PeMS04 /
PeMS08 ($24$ sensors $\times\,120$ five-minute bins, congested window).
Held-out fractions remove $f$ of the \emph{interior} sensors (endpoints
retained as boundaries), reseeded per run. All PINNs are $4\times64$
tanh MLPs trained with Adam ($10^{-3}$, $25$k steps). All models were
trained on a single Apple-silicon GPU (MPS backend); one $25$k-step
LC-PINN fit takes $\approx\!11$ minutes, while the EKF and
classical-kriging baselines run in seconds. 

%% file: sections/results_traffic.tex
\section{Results}
\label{sec:results}

We evaluate on four real freeway corridors with
independently-measured density --- two shock-rich trajectory corridors
(NGSIM US-101 and I-80, aggregated by Edie's definitions) and two
Caltrans loop-detector corridors (PeMS04, PeMS08; flow and speed) ---
holding out a fraction
of interior sensors and reporting the held-out rel-$L^2$ of
\eqref{eq:heldout-metric}. Unless noted, errors are means over five
seeds for every method; the fundamental diagram is Greenshields, calibrated per
corridor. We report four findings: physics-informed estimators
generalise to unmonitored sensors where data-driven models do not
(\S\ref{sec:results:generalise}); the advantage grows under sparse
sensing, sharply enough to produce a data-efficiency crossover
(\S\ref{sec:results:crossover}); the gain is attributable to the
conservation law specifically, not to smoothing
(\S\ref{sec:results:control}); and the loss-conditional network
recovers tuned-quality estimates while amortising the weight search
(\S\ref{sec:results:amortise}). An external PDEBench benchmark
corroborates the amortisation behaviour (\S\ref{sec:results:darcy}).

\subsection{Physics generalises to unmonitored sensors}
\label{sec:results:generalise}

\begin{figure}[t]
\centering
\includegraphics[width=0.82\linewidth]{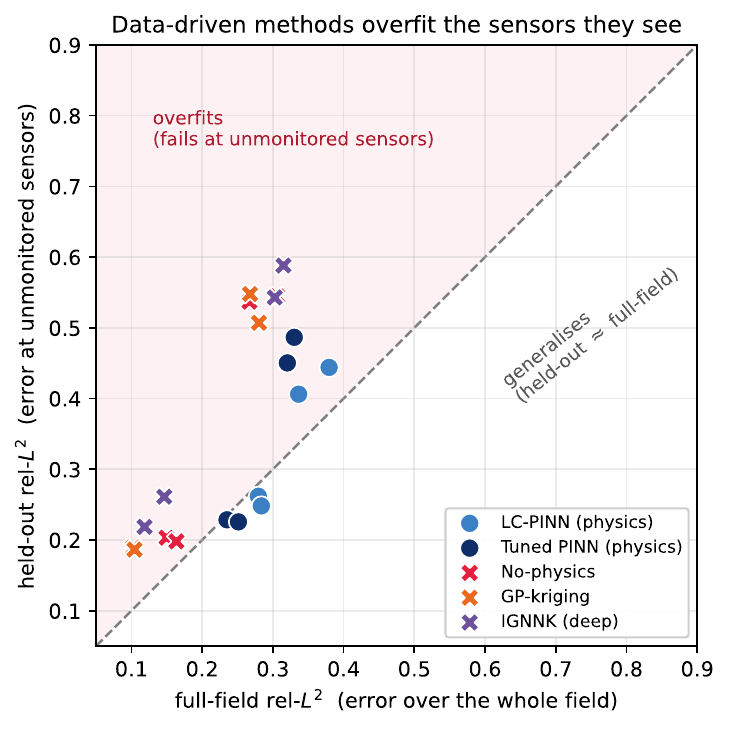}
\caption{Full-field vs.\ held-out error at $30\%$ hold-out, for the plotted methods on each corridor. Points on the diagonal generalise (held-out $=$
full-field); points above it overfit. Physics-informed estimators
(circles) lie on the diagonal; data-driven methods (crosses) sit in the
overfitting region.}
\label{fig:overfit}
\end{figure}

Table~\ref{tab:heldout-30} reports held-out error at a moderate
$30\%$ hold-out across all four corridors. On the PeMS corridors the
physics-informed estimators (LC-PINN, tuned) are strongest at held-out
sensors, although the no-physics network fits the \emph{observed} sensors
best. On the dense shock corridors
(NGSIM, I-80), data-driven kriging is best at this hold-out level; the
physics advantage there emerges only as sensing thins
(\S\ref{sec:results:crossover}). A trivial mean-density predictor
(Table~\ref{tab:heldout-30}, last row) anchors the scale: on the
shock-rich NGSIM corridor all estimators except the EKF sit well below
it, whereas on the near-free-flow PeMS corridors several baselines
(no-physics, GP/$k$NN-kriging, IGNNK, the EKF) fall
\emph{at or above} the floor on at least one corridor, while the
physics-informed estimators (LC-PINN and the tuned PINN) are the only
methods that clear it on all four. The pattern is
diagnostic: a purely data-driven estimator overfits the sensors it
sees and fails to extrapolate to the locations it does not.
Figure~\ref{fig:overfit} makes this visible across all methods and
corridors: plotting full-field error against held-out error, the
data-driven methods sit far above the diagonal (low full-field, high
held-out error), while the
physics-informed estimators lie on it (they generalise).

\begin{table}[t]
\centering
\caption{Held-out rel-$L^2$ at $30\%$ sensors held out (lower is
better; mean over seeds). Physics-informed estimators (LC-PINN,
tuned) lead at held-out on the PeMS corridors; on dense NGSIM
data-driven kriging is best, the crossover emerging only under sparsity
(\S\ref{sec:results:crossover}). Best per column in bold.}
\label{tab:heldout-30}
\begin{tabular}{lcccc}
\toprule
Method & NGSIM & I-80 & PeMS04 & PeMS08 \\
\midrule
LC-PINN (no weight search)   & 0.262 & 0.248 & \textbf{0.444} & \textbf{0.406} \\
Tuned LWR-PINN          & 0.229 & 0.226 & 0.487 & 0.450 \\
No-physics (data-only)  & 0.203 & 0.198 & 0.537 & 0.544 \\
GP-kriging              & \textbf{0.190} & 0.187 & 0.548 & 0.507 \\
$k$NN-kriging           & 0.192 & \textbf{0.182} & 0.582 & 0.486 \\
IGNNK (deep kriging)    & 0.219 & 0.261 & 0.588 & 0.543 \\
EKF (classical DA)      & 0.488 & 0.434 & 0.547 & 0.585 \\
\midrule
Mean predictor (floor)  & 0.371 & 0.437 & 0.545 & 0.485 \\
\bottomrule
\end{tabular}
\end{table}

\subsection{Data-efficiency under sparse sensing}
\label{sec:results:crossover}

\begin{figure*}[t]
\centering
\includegraphics[width=\textwidth]{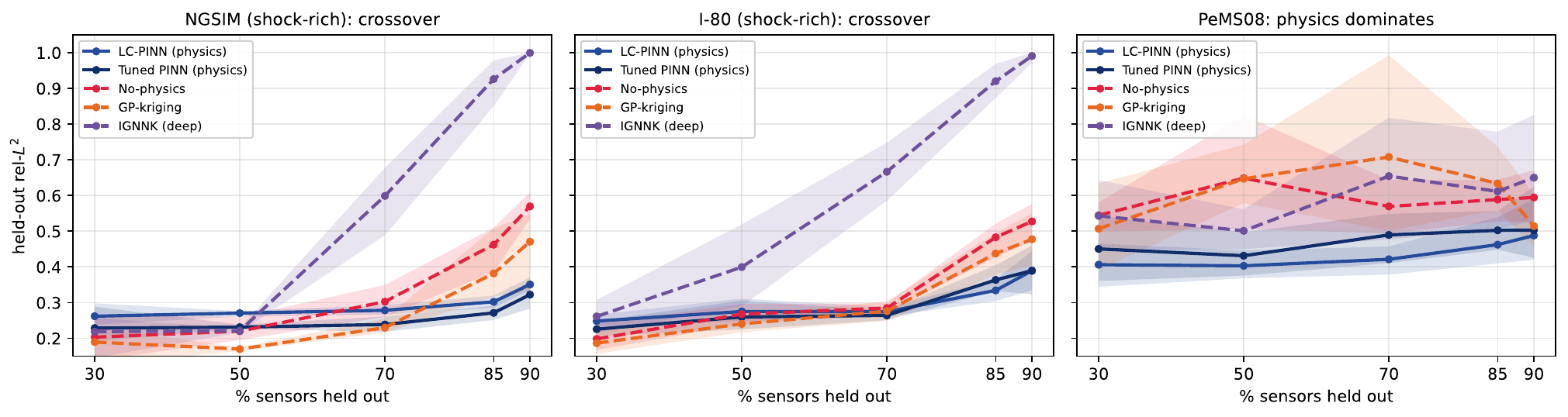}
\caption{Held-out rel-$L^2$ vs.\ fraction of sensors held out, on the two
shock-rich corridors NGSIM (left) and I-80 (centre) and the
near-free-flow PeMS08 (right). On \emph{both} shock corridors the
data-driven baselines (dashed) are best when dense but collapse as
sensors thin, while the physics-informed estimators (solid) stay flat
--- a clean data-efficiency crossover that replicates across two
independent freeways; on PeMS08 physics dominates throughout. Bands are
bootstrap 95\% CIs over seeds.}
\label{fig:crossover}
\end{figure*}

The defining result is how the methods behave as sensors thin.
Figure~\ref{fig:crossover} sweeps the held-out fraction from $30\%$ to
$90\%$ on the two shock corridors (NGSIM, I-80) and PeMS08. Most
data-driven baselines degrade sharply: on NGSIM the no-physics network
rises from $0.20$ to $0.57$, GP-kriging from $0.19$ to $0.47$, and the
deep IGNNK collapses from $0.22$ to $\approx\!1.0$; I-80 shows the same pattern (no-physics $0.20\!\to\!0.53$, GP-kriging
$0.19\!\to\!0.48$, IGNNK $0.26\!\to\!0.99$), and the PeMS corridors show
the same collapse (Table~\ref{tab:sweep}). The physics-informed
estimators instead stay nearly flat.

\begin{figure}[t]
\centering
\includegraphics[width=\linewidth]{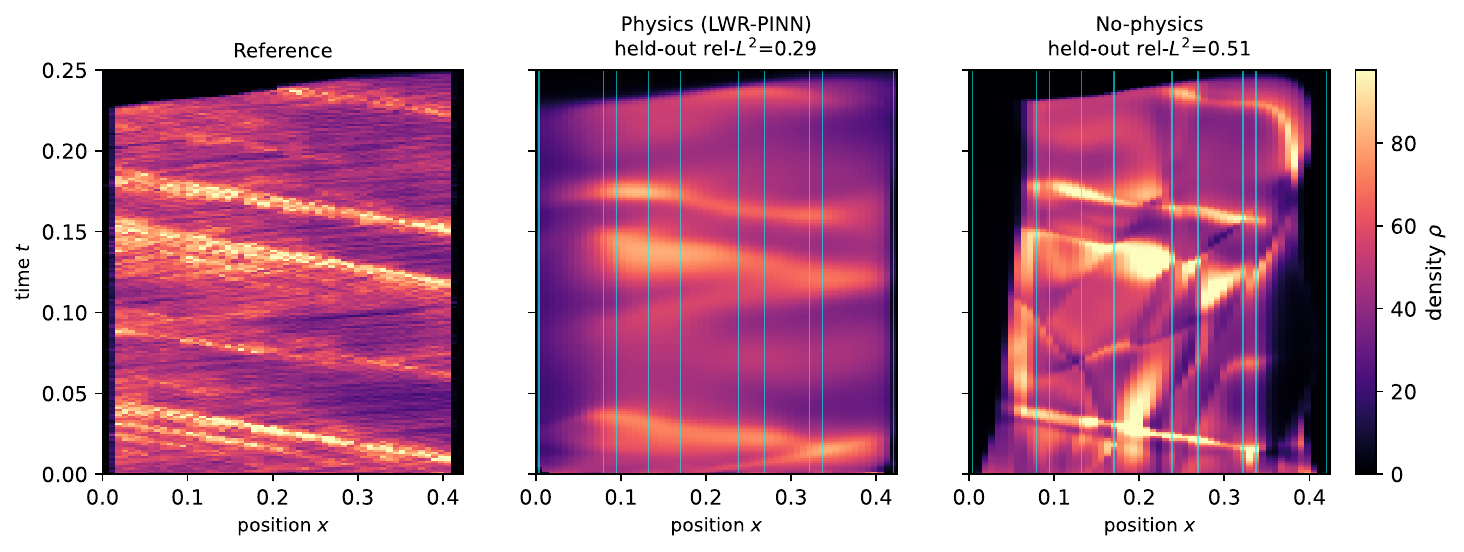}
\caption{Space-time density on NGSIM at $85\%$ of sensors held out:
reference (left), LWR-PINN (centre), no-physics (right); titles report
held-out rel-$L^2$ for the run shown. With only $15\%$ of sensors retained the
physics-informed estimator preserves the coherent backward-propagating
congestion structure, while the no-physics network fragments. The
physical prior smooths the finest stop-and-go waves --- the cost of
the prior --- but recovers the macroscopic state that data alone
cannot.}
\label{fig:field}
\end{figure}

Two regimes emerge. When sensors are dense, data-driven interpolation is
most accurate ($k$NN/GP $\approx\!0.18$--$0.19$ at $30\%$). As coverage
thins, the no-physics network, GP-kriging, and deep IGNNK \emph{collapse},
and on both shock corridors the LC-PINN overtakes them
by $\approx\!26$--$38\%$ relative \emph{to the no-physics network} at $85$--$90\%$
held out. The lone exception among data-driven methods is $k$NN
interpolation, which degrades gracefully: against $k$NN the margin is
smaller and emerges later ($\approx\!7$--$21\%$ at $85$--$90\%$) --- $k$NN
is most accurate when dense, the two are within seed noise at $85\%$, and
the LC-PINN pulls clearly ahead only at the sparsest $90\%$. The crossover is thus robust against
correlation-learning and deep baselines, and narrower --- but still
present on the shock corridors --- against simple local interpolation.
That it replicates across two independent freeways, not a single
corridor, is the central evidence that physics is the data-efficient
prior. On PeMS08 the LC-PINN is the most accurate method at \emph{every}
fraction ($\approx\!0.40$--$0.49$, ahead of the strongest classical
baseline at $\approx\!0.49$--$0.54$); on near-free-flow PeMS04 the LC-PINN
and the EKF trade places (LC-PINN best at $30$, $50$ and $85\%$, the EKF at
$70$ and $90\%$), as expected where little shock structure remains. A
seed-paired test on the shock corridors sharpens this: against $k$NN the
LC-PINN advantage is significant only at the sparsest $90\%$ (NGSIM
$+0.095$, $5/5$ seeds; I-80 $+0.043$, $4/5$; paired $95\%$ CIs excluding
zero) and within seed noise at $85\%$, whereas against the no-physics
network it is already large and significant by $85\%$ ($+0.16$ on NGSIM,
$+0.15$ on I-80; $5/5$ seeds) and against the collapsing deep baseline it
reaches $+0.6$ at $90\%$. On the shock corridors and PeMS08, the physics-informed estimators
degrade least as sensing becomes sparse.
Figure~\ref{fig:field} shows this qualitatively: at $85\%$ held out the
physics-informed reconstruction preserves the coherent stop-and-go
structure while the no-physics field fragments.

\begin{table}[t]
\centering
\caption{Sparse-sensor sweep, held-out rel-$L^2$ (mean over seeds).
Physics-informed error grows least on the shock corridors.}
\label{tab:sweep}
\setlength{\tabcolsep}{4pt}
\begin{tabular}{llccccc}
\toprule
& Method & .30 & .50 & .70 & .85 & .90 \\
\midrule
\multirow{6}{*}{\rotatebox{90}{NGSIM}}
 & LC-PINN      & 0.262 & 0.271 & 0.278 & 0.302 & 0.351 \\
 & No-physics   & 0.203 & 0.220 & 0.303 & 0.462 & 0.569 \\
 & GP-kriging   & 0.190 & 0.170 & 0.230 & 0.382 & 0.471 \\
 & $k$NN-krig   & 0.192 & 0.197 & 0.251 & 0.354 & 0.445 \\
 & IGNNK        & 0.219 & 0.221 & 0.599 & 0.926 & 0.999 \\
 & EKF (phys)   & 0.488 & 0.496 & 0.538 & 0.598 & 0.661 \\
\midrule
\multirow{6}{*}{\rotatebox{90}{I-80}}
 & LC-PINN      & 0.248 & 0.275 & 0.274 & 0.334 & 0.389 \\
 & No-physics   & 0.198 & 0.268 & 0.285 & 0.483 & 0.527 \\
 & GP-kriging   & 0.187 & 0.240 & 0.276 & 0.437 & 0.477 \\
 & $k$NN-krig   & 0.182 & 0.206 & 0.252 & 0.358 & 0.432 \\
 & IGNNK        & 0.261 & 0.399 & 0.666 & 0.920 & 0.991 \\
 & EKF (phys)   & 0.434 & 0.448 & 0.492 & 0.644 & 0.663 \\
\midrule
\multirow{6}{*}{\rotatebox{90}{PeMS04}}
 & LC-PINN      & 0.444 & 0.488 & 0.610 & 0.495 & 0.579 \\
 & No-physics   & 0.537 & 0.510 & 0.585 & 0.636 & 0.634 \\
 & GP-kriging   & 0.548 & 0.581 & 1.058 & 0.709 & 0.708 \\
 & $k$NN-krig   & 0.582 & 0.525 & 0.636 & 0.598 & 0.704 \\
 & IGNNK        & 0.588 & 0.543 & 0.606 & 0.826 & 0.988 \\
 & EKF (phys)   & 0.547 & 0.600 & 0.584 & 0.585 & 0.571 \\
\midrule
\multirow{6}{*}{\rotatebox{90}{PeMS08}}
 & LC-PINN      & 0.406 & 0.403 & 0.421 & 0.462 & 0.488 \\
 & No-physics   & 0.544 & 0.649 & 0.569 & 0.589 & 0.594 \\
 & GP-kriging   & 0.507 & 0.647 & 0.708 & 0.634 & 0.514 \\
 & $k$NN-krig   & 0.486 & 0.535 & 0.502 & 0.496 & 0.507 \\
 & IGNNK        & 0.543 & 0.501 & 0.654 & 0.611 & 0.650 \\
 & EKF (phys)   & 0.585 & 0.670 & 0.597 & 0.639 & 0.639 \\
\bottomrule
\end{tabular}
\end{table}

The deployment implication: on NGSIM the physics-informed estimator at
$15\%$ sensor coverage ($0.302$) matches the no-physics network at $30\%$
coverage ($0.303$), so it
reaches their accuracy with roughly half the sensors.

\subsection{Classical data assimilation is also outperformed}
\label{sec:results:ekf}

An extended Kalman filter (EKF) on the same calibrated LWR dynamics stays
bounded as sensors thin (NGSIM $0.49\!\to\!0.66$, PeMS08 $\approx\!0.60$)
instead of collapsing, so the conservation law resists sparsity however it
is imposed. On the corridors with shock structure, however, it is clearly
less accurate than the LWR-PINN (NGSIM $0.49$ vs.\ LC-PINN $0.26$ at
$30\%$, $0.66$ vs.\ $0.35$ at $90\%$;
Tables~\ref{tab:heldout-30},~\ref{tab:sweep}); only on near-free-flow
PeMS04 is it competitive. A soft, learnable physics prior fits the data
better than a rigid Godunov model, matching the ordering reported by Shi
et al.~\citep{shi2021lwr}. Held-out EKF error changes by under $0.05$
across a $3\times$ range of the observation-noise level, so the gap is not
an artifact of an undertuned filter.

\subsection{It is the physics, not smoothing}
\label{sec:results:control}

Replacing the LWR residual on NGSIM with a weight-matched
Laplacian-curvature penalty is worse than even the no-physics network at
every sparsity ($\approx\!0.54$--$0.60$, vs.\ LWR $0.23$--$0.32$): the
curvature penalty blurs the backward-propagating shock fronts that carry
the held-out signal. The gain is specific to the conservation law.

\subsection{Robustness checks}
\label{sec:results:robust}

\runinhead{Tuned baselines do not close the gap} Selecting the GP
length-scale and $k$NN neighbourhood per corridor on a validation split of
the \emph{observed} sensors moves their held-out error by only a few
points, in both directions. At $85$--$90\%$ held out the LC-PINN stays
ahead of the strongest tuned kriging on both shock corridors (NGSIM
$+0.06$/$+0.05$, I-80 $+0.03$/$+0.03$), though on NGSIM the margin falls
within seed noise ($4/5$ seeds; still significant on I-80 at $90\%$).
IGNNK collapses regardless of training budget. 

\runinhead{The crossover is not an FD-calibration artifact} The
fundamental diagram is fitted to aggregate flow--density data that
include held-out sensors. Recalibrating it on training sensors only leaves
NGSIM held-out error essentially unchanged (EKF $0.488\!\to\!0.487$ at
$30\%$, $0.661\!\to\!0.653$ at $90\%$; tuned PINN within $0.022$ at all
sparsities).

\subsection{Robustness to sensor noise}
\label{sec:results:noise}

With multiplicative noise $\tilde\rho = \rho\,(1+\sigma\epsilon)$,
$\epsilon\sim\mathcal{N}(0,1)$, scored against the clean field (NGSIM,
$50\%$ held out), the tuned physics-informed
network is nearly flat
($0.231\!\to\!0.235$ over $\sigma\in[0,0.5]$) while the no-physics network
degrades ($0.220\!\to\!0.367$), a $36\%$ relative gap at $\sigma{=}0.5$.

\subsection{Amortising the weight search}
\label{sec:results:amortise}

Fixed-weight PINNs require a per-corridor loss-weight search; the tuned
weights in the tables above were themselves found from an LC-PINN
sweep. LC-PINN removes that search: a single training run covers the
whole weight frontier. On NGSIM, one LC-PINN training (held-out
$0.262$) matches --- indeed slightly beats --- the best of a measured
$20$-point random-$\lambda$ search (best $0.267$, median $0.380$) at
$\approx\!5\times$ less compute and without the gamble; relative to an exhaustive grid the
saving is $18$--$58\times$. On the PeMS
corridors LC-PINN goes further, slightly outperforming the tuned
fixed-weight PINN (Table~\ref{tab:heldout-30}), so on these corridors
the amortised estimate is also slightly more accurate. Consistent with
Proposition~\ref{prop:invariance}, $K$-shot averaging reduces variance
but does not change the mean; we
report centre-$\lambda$ estimates throughout.

\begin{figure}[t]
\centering
\includegraphics[width=0.86\linewidth]{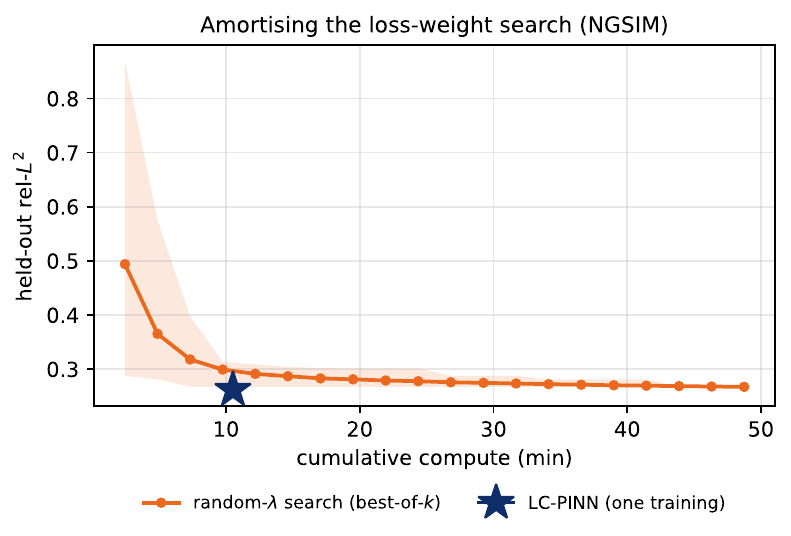}
\caption{Amortising the loss-weight search on NGSIM. The
random-$\lambda$ search of fixed-weight PINNs improves with cumulative
compute but with high variance (shaded, 10--90\%); one LC-PINN training
($\star$) matches the search's best accuracy at $\sim\!5\times$ less
compute and without the gamble.}
\label{fig:amort}
\end{figure}

\subsection{External benchmark: PDEBench Darcy}
\label{sec:results:darcy}

To test the amortisation behaviour outside traffic, we apply the same
conditioning to the PDEBench 2D Darcy-flow benchmark. On a \emph{single} Darcy instance a
fixed-weight PINN is more accurate (rel-$L^2$ $0.33$ vs.\ LC-PINN
$0.38$), the cost expected when one model covers a family instead of
fitting one instance. Strong-form PINNs
are a known hard case on Darcy's rough permeability field, so absolute
accuracy is modest for all variants; the value of the comparison is
that the amortisation--accuracy trade-off observed in traffic
reproduces on an independent, public benchmark.

%% file: sections/theory_traffic.tex
\section{Why Physics Generalises: Theory}
\label{sec:theory}

We give a theoretical account of the empirical asymmetry --- physics
generalises to unmonitored sensors, data-driven methods do not. We give two results: an identifiability statement in the inviscid limit,
and an error certificate for the viscous-regularised model we train
($N_\varepsilon[u]=\partial_t u+\partial_x Q(u)-\varepsilon\,\partial_{xx}u$,
$\varepsilon=10^{-2}$, with exact density $\rho$ and network residual
$R_\theta=N_\varepsilon[u_\theta]$).

\runinhead{Sparse sensors suffice (identifiability)}
In the inviscid smooth (pre-shock) limit, the conservation law transports
each observation along characteristics:

\begin{proposition}[Identifiability]\label{prop:ident}
Let $\rho\in C^1$ solve $\partial_t\rho+\partial_x Q(\rho)=0$ on a
shock-free region. Along each characteristic $\dot x=Q'(\rho)$, $\rho$ is
constant; hence for any $(x,t)$ whose characteristic meets a sensor line
$\ell_i=\{x_i\}\times(0,T)$ at $(x_i,s)$, $\rho(x,t)=\rho(x_i,s)$ is
determined. The field is thus identified on the union of the sensors'
domains of determinacy.
\end{proposition}

\noindent Each sensor line determines a two-dimensional region, not just
itself, because the PDE \emph{transports} the observation to unobserved
$(x,t)$. A correlation-based interpolant has no transport --- an
observation constrains only its own location --- which is the structural
origin of the data-efficiency gap (a single sensor informs an area, not a
point). Outside the domain of determinacy the field is recoverable by no
method; this limit applies to every method.

\runinhead{A certificate at unmonitored locations}
For the viscous problem there is a computable a-posteriori bound.

\begin{theorem}[Held-out error certificate]\label{thm:cert}
Let $\varepsilon>0$ and $\rho,u_\theta\in C^{2,1}(\overline\Omega)$. Then
\begin{equation}
\norm{u_\theta-\rho}_{L^2(\Omega)}^2
\;\le\; C\big(\norm{R_\theta}_{L^2(\Omega)}^2 + \text{data misfit}\big),
\end{equation}
with $C=C(T,\varepsilon,\sup|Q'|)$ growing at most like $e^{cT}$.
\end{theorem}

\noindent The point is \emph{which} quantities bound the error. The
residual $\norm{R_\theta}_{L^2(\Omega)}$ is an interior term on collocation
points that carry \emph{no labels} and can be placed densely throughout
$\Omega$. So a small training residual plus a small data misfit certifies
small error over the \emph{whole} domain --- \emph{including at held-out,
sensor-free locations}. A data-driven estimator minimises only the data
misfit at observed sensors and has no residual term: nothing it minimises
controls its error away from those sensors. This asymmetry is the theoretical counterpart of the data-efficiency
effect that Figure~\ref{fig:overfit} shows.

\runinhead{Loss-conditioning and $K$-shot}
Finally, Proposition~\ref{prop:invariance} (Section~\ref{sec:method:inference})
states that at the optimum the loss weights $\lambda$ are inert; at finite training, $K$-shot averaging reduces
variance but leaves bias unchanged, which is why it reduces spread but not
the mean.

\runinhead{Scope}
Proposition~\ref{prop:ident} is inviscid and pre-shock; the certificate
(Theorem~\ref{thm:cert}) needs no such restriction, so the two bracket the
physical small-$\varepsilon$ regime. The certificate is a-posteriori (it
bounds the error \emph{given} an achieved residual; it does not assert one
is achievable), and its constant degrades for long horizons and sharp
fronts. In practice the residual is evaluated on finitely many
collocation points, which estimate the continuous norm
$\norm{R_\theta}_{L^2(\Omega)}$. Denser collocation makes this estimate
more reliable, so the certificate can be trusted more; it does not make
the bound itself tighter, because $C$ depends on the horizon, the
viscosity and the flux, not on the number of points. Near shocks the
residual concentrates in thin regions that sparse uniform collocation
can miss, which is where denser or adaptive sampling matters most. We state it for boundary and initial data; it extends to interior
sensors by treating each sensor line as an internal boundary.

%% file: sections/discussion_traffic.tex
\section{Discussion and Limitations}
\label{sec:discussion}

\runinhead{What the result is, and is not}
The claim is \emph{relative}: under sparse sensing, a conservation-law
prior generalises to unmonitored locations where data-driven estimators
collapse. Absolute held-out errors remain high (LC-PINN rel-$L^2$ $0.26$--$0.35$
on NGSIM, $0.40$--$0.61$ on the PeMS corridors); estimating the state at
locations with no sensor is intrinsically hard, and we do not claim
deployable absolute accuracy. Whether these levels are useful depends on
the downstream task. Coarse tasks, such as detecting congestion or
classifying the level of service, need the location and extent of
congested regions more than exact density values, and the estimator
recovers this macroscopic structure. Fine tasks, such as travel-time
prediction or ramp-metering control, need tighter accuracy than the
sparsest settings provide. Under extreme sparsity the estimate is best
read as a structural reconstruction, not as a substitute for a sensor. We also do not claim that LC-PINN beats a
\emph{tuned} fixed-weight PINN at a single weighting: widening the LC-PINN to $15\times$ the parameters narrows but does not
close its gap to the tuned PINN on NGSIM, and on PDEBench Darcy a fixed-weight PINN is more accurate on a
single instance. LC-PINN's value is amortisation (no per-corridor weight
search) and, on the PeMS corridors, a modest accuracy gain. Finally, the
\emph{direction} of the NGSIM crossover --- data-driven kriging ahead at
the densest sensing --- is a property of our held-out-only metric on the two
shock corridors; we report it as an observation, not as a contradiction of
prior full-field studies \citep{shi2021lwr} in which the physical prior
is uniformly ahead.

\runinhead{Where the advantage is strongest}
The crossover is sharpest on the shock-rich NGSIM corridor, where the
conservation law is most constraining. On the near-free-flow PeMS
corridors physics still leads on PeMS08 and is closer on PeMS04,
and classical $k$NN-kriging remains competitive on PeMS08
($\approx\!0.49$--$0.54$): with
little shock structure to exploit, the physics prior is less decisive.
The conservation law helps most where the traffic has shock structure.

\runinhead{Modelling choices and caveats}
We use the first-order LWR model with a per-corridor calibrated
Greenshields fundamental diagram; results are similar with a triangular
diagram. Jointly estimating the FD parameters
($\rho_{\max}, v_f, \varepsilon$) is orthogonal to the data-efficiency question we study and is left
aside. All results concern single, continuous 1-D corridors. Urban networks
with signalised intersections, on-ramps and off-ramps add junction
conditions and source terms that the single-corridor LWR model does not
represent, so our results should not be read as evidence for such
networks. The conditioning itself is not 1-D-specific (it transfers to
the 2-D Darcy benchmark, \S\ref{sec:results:darcy}), but a network-level
traffic estimator with junction models is future work.
The
four corridors all carry independently-measured density (trajectories,
or flow and speed); we deliberately exclude speed-only corridors (e.g.\
PeMS-BAY, METR-LA) because their density must be inferred from the same
fundamental diagram the physics enforces, which would confound the
comparison.
L-BFGS finishing, which helps on the synthetic PDEs of the conditioning
literature, does not help on traffic LWR; we train with Adam only. The
deep kriging baseline (IGNNK) is designed for large 2-D sensor
graphs and are out of their comfort zone on a short 1-D corridor; we
report them with this caveat rather than as a rigged comparison.
Qualitatively, the estimator recovers macroscopic congestion structure
but smooths the finest stop-and-go waves.

\runinhead{Future work}
Second-order traffic models (ARZ) would give a richer prior on the PeMS
corridors, and the conditioning extends naturally to a network-level
estimator over many corridors, with the loss weights amortised across
them.